\documentclass[11pt,a4paper]{article}
\usepackage[hyperref]{acl2019}
\usepackage{times}
\usepackage{latexsym}
\usepackage{array}
\usepackage{graphicx}
\usepackage{multirow}
\usepackage{color}
\usepackage{relsize}

\usepackage{url}

\aclfinalcopy
\def\aclpaperid{164} 

\title{Fact or Factitious? Contextualized Opinion Spam Detection} 

\author{Stefan Kennedy, Niall Walsh\thanks{Joint first author with Stefan Kennedy}, Kirils Sloka, Andrew McCarren and Jennifer Foster \\
   School of Computing \\
   Dublin City University\\
  \texttt{\smaller \{\texttt{15902803,15384141,13405888\}@mail.dcu.ie}} \\
   \texttt{\smaller \texttt{\{Andrew McCarren,Jennifer.Foster\}@dcu.ie}} \\\\
   \\}
  
\date{}

\begin{document}
\maketitle
\begin{abstract}
In recent years, it has been shown that falsification of online reviews can have a substantial, quantifiable effect on the success of the subject. This creates a large enticement for sellers to participate in review deception to boost their own success, or hinder the competition.
Most current efforts to detect review deception are based on supervised classifiers trained on syntactic and lexical patterns. However, recent neural approaches to classification have been shown to match or outperform state-of-the-art methods.
In this paper, we perform an analytic comparison of these methods, and introduce our own results. By fine-tuning Google's recently published transformer-based architecture, BERT, on the fake review detection task, we demonstrate near
state-of-the-art performance,
achieving over 90\% accuracy on a widely used deception detection dataset.
\end{abstract}

\section{Introduction}
Online reviews of products and services have become significantly more important over the last two decades. Reviews influence customer purchasing decisions through review score and volume of reviews~\cite{MASLOWSKA20171}. 
It is estimated that as many as 90\% of consumers read reviews before a purchase~\cite{doi:10.1080/07421222.2018.1440758}
and that the conversion rate of a product increases by up to 270\% as it gains reviews. For high price products, reviews can increase conversion rate by 380\%~\cite{Askalidis16}. 

With the rise of consumer reviews comes the problem of deceptive reviews.  It has been shown that in competitive, ranked conditions it is worthwhile for unlawful merchants to create fake reviews. For TripAdvisor, in 80\% of cases, a hotel could become more visible than another hotel using just 50 deceptive reviews~\cite{lappas2016}. Fake reviews are an established problem -- 20\% of Yelp reviews are marked as fake by Yelp's algorithm \cite{Luca2016FakeIT}.


First introduced by \newcite{Jindal:2007:RSD:1242572.1242759}, the problem of fake review detection has been tackled from the perspectives of opinion spam detection and deception detection. It is usually treated as a binary classification problem using traditional text classification features such as word and part-of-speech n-grams, structural features obtained from syntactic parsing~\cite{feng-etal-2012-syntactic}, topic models~\cite{Hernandez-Castaneda:2017:CDD:3049504.3049635}, psycho-linguistic features obtained using the \textit{Linguistic Inquiry and Word Count}~\cite{ott-etal-2011-finding,Hernandez-Castaneda:2017:CDD:3049504.3049635, pennebaker2015development} and non-verbal features related to reviewer behaviour~\cite{you-etal-2018-attribute, Wang2017FakeRD,aghakhani2018, DBLP:journals/corr/abs-1903-08289}

We revisit the problem of fake review detection by comparing the performance of a variety of neural and non-neural approaches on two freely available datasets, a small set of hotel reviews where the deceptive subset has been obtained via crowdsourcing~\cite{ott-etal-2011-finding} and a much larger set of Yelp reviews obtained automatically~\cite{Rayana:2015:COS:2783258.2783370}. We find that features based on reviewer characteristics can be used to boost the accuracy of a strong bag-of-words baseline. We also find that neural approaches perform at about the same level as the traditional non-neural ones. Perhaps counter-intuitively, the use of pretrained non-contextual word embeddings do not tend to lead to improved performance in most of our experiments. However, our best performance is achieved by fine-tuning BERT embeddings~\cite{devlin2018bert} on this task. On the hotel review dataset, bootstrap validation accuracy is 90.5\%, just behind the 91.2\% reported by  \newcite{feng-etal-2012-syntactic} who combine bag-of-words with constituency tree fragments.

\section{Data}
Collecting data for classifying opinion spam is difficult because human labelling is only slightly better than random \cite{ott-etal-2011-finding}. Thus, it is difficult to find large-scale ground truth data. We experiment with two datasets:
\begin{itemize}
    \item \texttt{OpSpam}~\cite{ott-etal-2011-finding}: This dataset contains 800 gold-standard, labelled reviews.  These reviews are all deceptive and were written by paid, crowd-funded workers for popular Chicago hotels. Additionally this dataset contains 800 reviews considered truthful, that were mined from various online review communities. These truthful reviews cannot be considered gold-standard, but are considered to have a reasonably low deception rate. 
    \item \texttt{Yelp}~\cite{Rayana:2015:COS:2783258.2783370}: This is the largest ground truth, deceptively labelled dataset available to date. The deceptive reviews in this dataset are those that were filtered by Yelp's review software for being manufactured, solicited or malicious. Yelp acknowledges that their recommendation software makes errors\footnote{\label{note1}https://www.yelpblog.com/2010/03/yelp-review-filter-explained}. Yelp removed 7\% of its reviews and marked 22\% as not recommended\footnote{https://www.yelp.com/factsheet}. This dataset is broken into three review sets, one containing 67,395 hotel and restaurant reviews from Chicago, one containing 359,052 restaurant reviews from NYC and a final one containing 608,598 restaurant reviews from a number of zip codes. There is overlap between the zip code dataset and the NYC dataset, and it is known that there are significant differences between product review categories~\cite{blitzer2007biographies} (hotels and restaurants) so we will only use the zip code dataset in training our models. Due to the memory restrictions of using convolutional networks, we filter the reviews with an additional constraint of being shorter than 321 words. This reduces the size of our final dataset by 2.63\%. There are many more genuine reviews than deceptive, so we extract 78,346 each of genuine and deceptive classes to create a balanced dataset. The entire dataset contains 451,906 unused reviews.
\end{itemize}
\section{Methods}
We train several models to distinguish between fake and genuine reviews. The non-neural of these are logistic regression, and support vector machines \cite{Cortes1995}, and the
neural are feed-forward networks, convolutional networks and long short-term memory networks~\cite{LeCun:1998:CNI:303568.303704, Jacovi2018UnderstandingCN, Hochreiter:1997:LSM:1246443.1246450}. We experiment with simple bag-of-word input representations and, for the neural approaches, we also use pre-trained word2vec embeddings~\cite{DBLP:conf/icml/LeM14}. In contrast with word2vec vectors which provide the same vector for a particular word regardless of its sentential context, we also experiment with contextualised vectors. Specifically, we utilize the BERT model developed by Google \cite{devlin2018bert} for fine-tuning pretrained  representations.

\section{Experiments}
\subsection{Feature Engineering}
\begin{figure*}[ht]
\includegraphics[width=16cm]{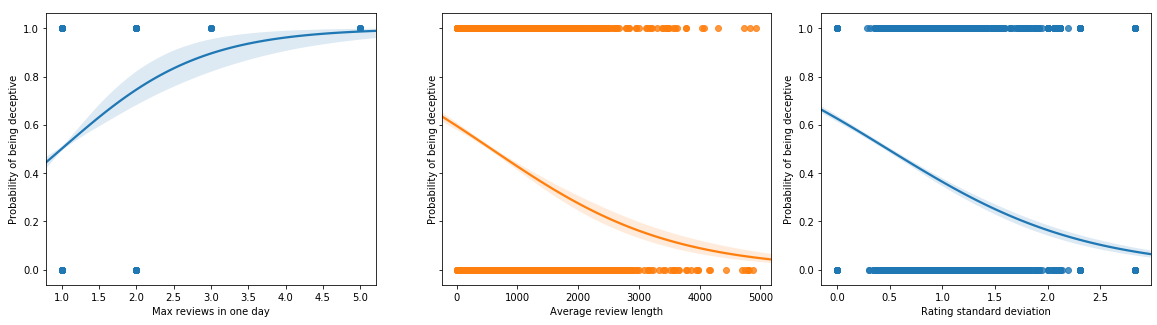}
\caption{Separability of three most significant metadata features. Max reviews in one day is computed over randomly sampled, equal numbers of each class. The vertical axis represents deceptive as 1.0 and genuine as 0.0. }
\centering
\end{figure*}
Following \newcite{Wang2017FakeRD}, we experiment with a number of features on the Yelp dataset:
\begin{itemize}
  \item Structural features including review length, average word and sentence length, percentage of capitalized words and percentage of numerals.
  \item Reviewer features including maximum review count in one day, average review length, standard deviation of ratings, and percentage of positive and negative ratings.
  \item Part-of-Speech (POS) tags as percentages.
  \item Positive and negative word sentiment as percentages.
\end{itemize}
Feature selection using logistic regression found that some features were not predictive of deception. In particular POS tag percentages and sentiment percentages were not predictive. Metadata about the author of the review was the most predictive of deception, and the highest classification performance occurred when including only reviewer features in conjunction with bag-of-word vectors. Separation of these features displayed in Figure 1 shows that a large number (greater than 2) of reviews in one day indicates that a reviewer is deceptive. Conversely a long (greater than 1000) average character length of reviews is indicative that a reviewer is genuine. The standard deviation of a user's ratings is also included as a large deviation is an indicator of a review being genuine. 
For the remainder of experiments, we concatenate the word representations with a scaled (from 0 to 1) version of these user features. Note that reviewer features are not available for the OpSpam dataset.

\subsection{Experimental Details}
\subsubsection*{Evaluation} For the smaller OpSpam dataset we report results with both 5-Fold cross validation and bootstrap validation repeated 10 times. For small datasets and a small number of splits, K-Fold is known to be subject to high variance. Additionally bootstrap validation is known to be subject to high bias in some contexts \cite{Kohavi:1995:SCB:1643031.1643047}. We therefore report results for both forms of validation. In all forms of validation we create stratified train and test sets.
For the larger Yelp dataset we use the balanced set described in section 2. As this dataset is substantially large enough we use 10-Fold cross validation only to obtain results. 

\subsubsection*{Non-Neural Models} For both OpSpam and Yelp datasets we design our models with similar methods. In the logistic regression and SVM experiments, words are represented in TF-IDF format, and in the case of Yelp only the most relevant 10,000 words are represented.
Repeated experiments found that both linear and non-linear SVM kernels produced comparable performance. Applying grid search with the Yelp dataset found that a linear kernel could reach the highest accuracy. 

\subsubsection*{Neural Models} For the Yelp dataset, neural classifiers use early stopping with a patience of 6 epochs of waiting for an improvement in validation loss. The same filtered, balanced dataset is used as input to all classifiers, and we use a hold out set of 1000 samples (6.38\% of the balanced data) to verify performance.

\paragraph{Word2vec} We use word2vec embeddings pretrained with a dimensionality of 300 on a Google News dataset\footnote{https://code.google.com/archive/p/word2vec/}. This model was pretrained using a skip-gram architecture.

\paragraph{FFNNs} We model FFNNs using a network containing two hidden dense layers. For both layers we use ReLU activation and l2 regularization, and we use sigmoid activation on the output layer. For the Yelp data, user features are directly concatenated to the BoW representation. For word2vec embeddings, the embeddings are first flattened to a single dimension before concatenation. The model used for OpSpam contains 32 units in the first hidden layer, and 16 units in the second. The model used for Yelp contains 16 units in the first hidden layer, and 8 units in the second. Models for both datasets use a dropout rate of 0.25 between the two hidden layers.

\paragraph{CNNs} Convolutional networks are modelled in different ways for BoW and word2vec embedding representations. As BoW is represented in a single dimension, we create a convolutional layer with a kernel height of 1 unit and width of 10 units. This kernel slides horizontally along the BoW vector. For word2vec embeddings we position word vectors vertically in the order they occur, as has been implemented in earlier research \cite{kim-2014-convolutional}. In this case the kernel has a width equal to the dimensionality of the word vectors and slides vertically along the word axis. We use a kernel height of 10, containing 10 words in each kernel position. Both BoW and word2vec embedding models use 50 filters. Following the convolutions the result is passed through a pooling layer, and a dropout rate of 0.5 is applied before the result is flattened. In the case of Yelp this flattened result is concatenated with the user features of the review. Two hidden dense layers follow this, both using ReLU activation and l2 regularization. Both hidden layers contain 8 units and are followed by an output layer that uses sigmoid activation. For the OpSpam dataset, the BoW model uses a pool size of (1, 10) and the word2vec embedding implementation uses a pool size of (5, 1). For the Yelp dataset both BoW and word2vec embedding models use global max pooling.

\paragraph{LSTMs} In the implementation of LSTMs, models for both BoW and word2vec embeddings directly input word representations to an LSTM layer. Numerous repeated runs with different numbers of LSTM layers and units found that the optimal accuracy occurs at just one layer of 10 units. We model implementations for both OpSpam and Yelp datasets using this number of layers and units. In the case of the Yelp dataset, the output of the LSTM layer is concatenated with user features. This is followed by 2 hidden dense layers using ReLU activation and l2 regularization, each containing 8 units, followed by an output layer using sigmoid activation.

\paragraph{BERT} 
 We fine-tune the \texttt{bert-base-uncased model} on the OpSpam dataset and perform stratified validation using both 5-Fold validation and bootstrap validation repeated 10 times. For fine-tuning we use a learning rate of 2e-5, batch size of 16 and 3 training epochs.

Two implementations of fine-tuning are used to verify results. One implementation is the BERT implementation published by Google alongside the pretrained models, and the other uses the `op-for-op' reimplementation of BERT created by Hugging Face\footnote{https://github.com/huggingface/pytorch-pretrained-BERT}.

\subsection{Results}
The results of performing validation on these models are shown in Tables 1, 2 and 3.
Table 1 shows that SVMs slightly
outperform logistic regression, and that the Yelp dataset represents a much harder challenge than the OpSpam one.

Contrary to expectations, Table 2 shows that pretrained word2vec embeddings do not improve performance, and in the case of OpSpam BoW can substantially outperform them. We do not yet know why this might be case.

The BERT results in Table 3 show that the Google TensorFlow implementation performs substantially better than PyTorch in our case. This is an unexpected result and more research needs to be carried out to understand the differences. We also report that Google's TensorFlow implementation outperforms all other classifiers tested on the OpSpam dataset, providing tentative evidence of contextualized embeddings outperforming all non-contextual pre-trained word2vec embeddings and BoW approaches.
\begin{table*}[t!]
\centering
\begin{tabular}{| c | c  c | c |}
  \hline
  & \multicolumn{2}{c|}{\textbf{OpSpam}} & \textbf{Yelp}\\
  & 5-Fold & Bootstrap & 10-Fold\\
  \hline
  Logistic Reg & 0.856 & 0.869 & 0.713\\
  \hline
  SVM & 0.864 & 0.882 & 0.721\\ 
  \hline
\end{tabular}
\caption{Non-neural Classifier Accuracy}
\end{table*}
\begin{table*}[t!]
\centering
\begin{tabular}{| c | c | c | c | c || c | c|}
  \hline
  & \multicolumn{4}{c||}{\textbf{OpSpam}} &  \multicolumn{2}{c|}{\textbf{Yelp}}\\
  & \multicolumn{2}{c|}{\textit{BoW}} & \multicolumn{2}{c||}{\textit{word2vec}} & \textit{BoW} & \textit{word2vec}\\
  & 5-Fold & Bootstrap & 5-Fold & Bootstrap & \multicolumn{2}{c|}{}  \\
  \hline
  FFNN & 0.888 & 0.883 & 0.587 & 0.605 & 0.708 & 0.704\\
  \hline
  CNN & 0.669 & 0.639 & 0.800 & 0.822 & 0.722 & 0.731\\
  \hline
  LSTM & 0.876 & 0.876 & 0.761 & 0.769 & 0.731 & 0.727 \\
  \hline
\end{tabular}
\quad
\caption{Neural classifier accuracy using bag-of-words (BoW) and non-contextual (word2vec) word embeddings}
\end{table*}

\begin{table}[h!]
\begin{center}
\begin{tabular}{ | m{5em} | m{5em}| m{4em} | } 
\hline
& TensorFlow & PyTorch \\
\hline
 K-Fold & 0.891 & 0.862 \\ 
\hline
 Bootstrap & 0.905 & 0.867 \\ 
\hline
\end{tabular}
\caption{Accuracy performance of BERT implementations in TensorFlow and PyTorch (OpSpam)}
\end{center}
\end{table}

By inspecting the results of evaluation on a single 5-Fold test set split for the BERT experiments, we see that there are an approximately equal number of false negatives (15), and false positives (14). There appears to be a slight tendency for the model to perform better when individual sentences are longer, and when the review is long. In the case of our 29 incorrect classifications the number of words in a sentence was 16.0 words, compared to 18.4 for correct classifications. Entire reviews tend to be longer in correct classifications with an average length of 149.0 words, compared to 117.6 for incorrect classifications. Meanwhile the average word length is approximately 4.25 for both correct and incorrect classifications.

\section{Application of Research}
We have developed a frontend which retrieves business information from Yelp and utilizes our models to analyze reviews. Results are displayed in an engaging fashion using data visualization and explanations of our prediction. We display a deception distribution of all reviews for the product. This includes how many reviews are classified as deceptive or genuine, shown in buckets at 10\% intervals of confidence. This allows users to quickly determine if the distribution is different to a typical, expected one. This tool also enables users to view frequency and average rating of reviews over time. This information can be used to spot unusual behaviour at a given time, such as a sudden increase in activity, where that activity is creating a positive or negative rating score.
 The aim of this web application is to highlight the ability of our models to detect fake reviews, and allows interactions that drill down on specific details such as the impact of individual words on the overall evaluation. Additional features enrich the evaluation by performing statistical analysis on the users who wrote the retrieved reviews. We use badges to show the significance of this analysis, where a badge is given to show a deceptive or genuine indicator. Reviews can receive badges for the user's average review length, standard deviation of review scores and maximum number of reviews in one day. This adds a layer of transparency to the data, allowing us to give a more informative verdict on the review itself.
\begin{figure}[h]
\includegraphics[width=7cm]{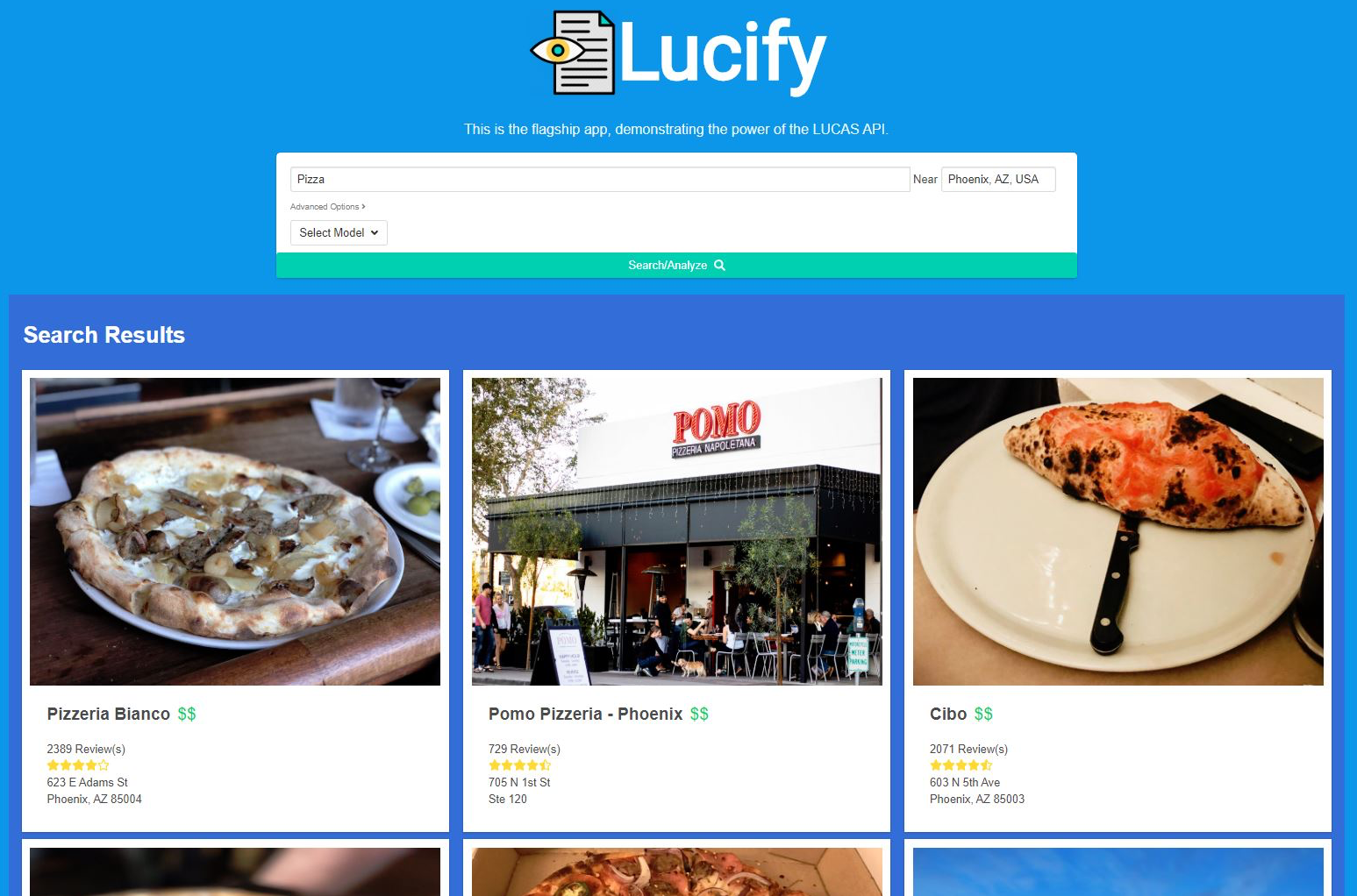}
\caption{Search page of web interface.}
\centering
\end{figure}

The models developed in this research are publicized through our API. 
The web application provides an option to set the model used in requests, providing easy access to experimentation. This is an open-source\footnote{https://github.com/CPSSD/LUCAS} project implemented in the React\footnote{https://reactjs.org} Javascript web interface library and Flask\footnote{http://flask.pocoo.org} Python server library respectively.

\begin{figure}[h]
\includegraphics[width=7cm]{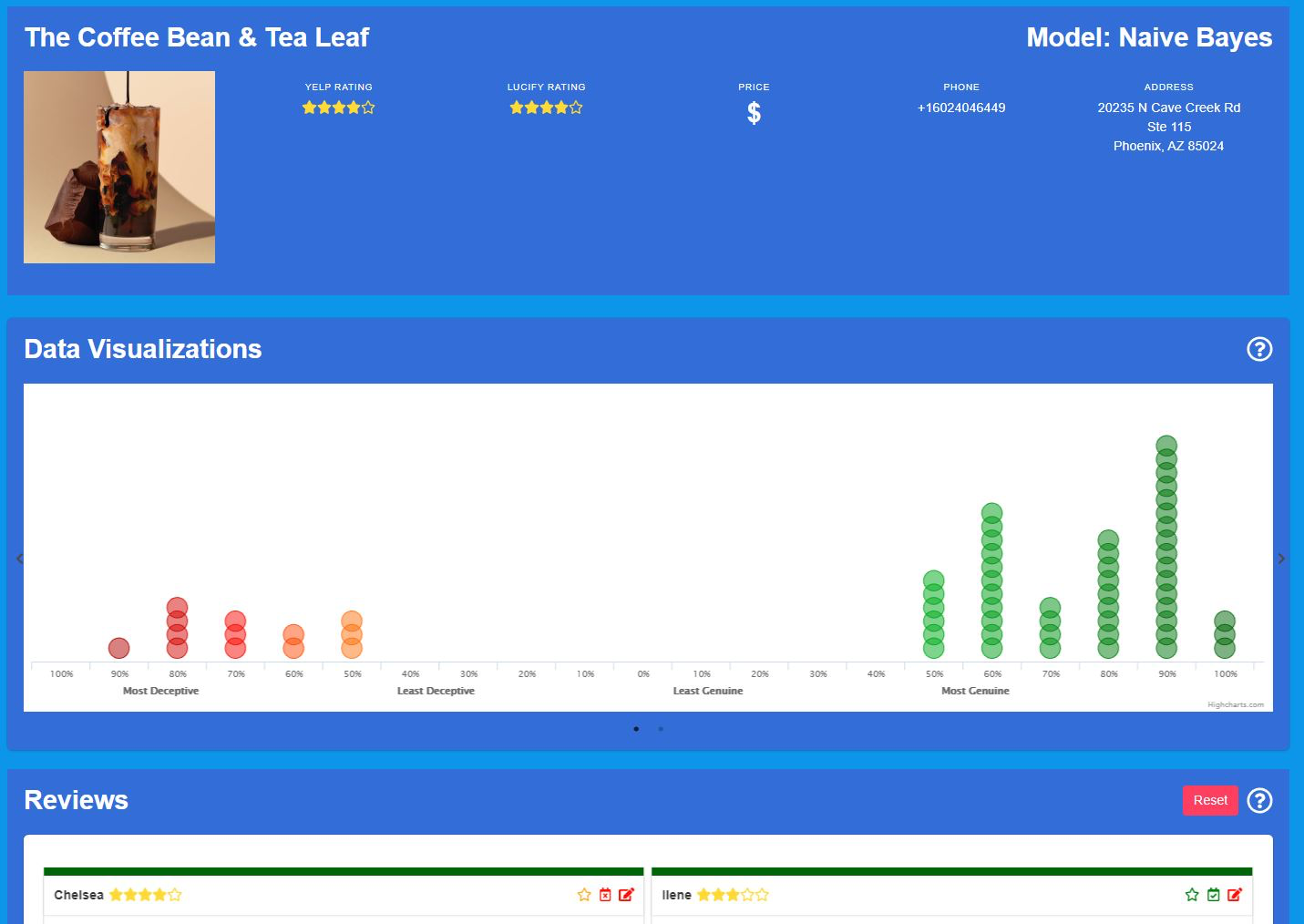}
\caption{Sample Visualization of Reviews}
\centering
\end{figure}

\section{Conclusion}
We have conducted a series of classification experiments on two freely available deceptive review datasets. The dataset created by crowd-sourcing deceptive reviews results in an easier task than the real-world, potentially noisy, dataset produced by Yelp. On the Yelp dataset, we find that features that encode reviewer behaviour are important in both a neural and non-neural setting. The best performance on the OpSpam dataset, which is competitive with the state-of-the-art, is achieved by fine-tuning with BERT.
Future work involves understanding the relatively poor performance of the pretrained non-contextual embeddings, and experimenting with conditional, more efficient generative adversarial networks.

 \section*{Acknowledgments} We thank the reviewers for their helpful comments. The work presented in this paper is the  final year project of the first three authors, conducted as part of their undergraduate degree in Computational Problem Solving and Software Development in Dublin City University.
 We thank the  ADAPT  Centre  for  Digital  Content  Technology for providing computing resources. The  ADAPT  Centre  for  Digital  Content  Technology  is  funded  under  the  Science Foundation Ireland Research Centres Programme (Grant 13/RC/2106) and is co-funded under the European Regional Development Fund.

\bibliography{acl2019}
\bibliographystyle{acl_natbib}



\end{document}